\documentclass[
twocolumn,
]{ceurart}

\usepackage{listings}
\usepackage{booktabs}
\usepackage{multirow}
\usepackage{makecell}
\usepackage{arydshln}
\usepackage{pifont}
\usepackage{dashrule}
\usepackage{amsmath}
\usepackage{enumitem}
\usepackage{multicol}
\usepackage[ruled, lined, linesnumbered, commentsnumbered, noend]{algorithm2e}

\SetCommentSty{mycommfont}

\definecolor{amethyst}{rgb}{0.6, 0.4, 0.8}
\begin{document}

\copyrightyear{2023}
\copyrightclause{Copyright for this paper by its authors. Use permitted under Creative Commons License Attribution 4.0 International (CC BY 4.0).}
\conference{Fifth Knowledge-aware and Conversational Recommender Systems (KaRS) Workshop @ RecSys 2023, September 18--22 2023, Singapore.}

\title{Extraction of Atypical Aspects from Customer Reviews: Datasets and Experiments with Language Models}


\author[1]{Smita Nannaware}[%
orcid=0009-0002-7185-7186,
email=snannawa@charlotte.edu,
]
\address[1]{Department of Computer Science, University of North Carolina at Charlotte, Charlotte, NC 28223, USA}

\author[1]{Erfan Al-Hossami}[%
orcid=0000-0002-8436-8974,
email=ealhossa@charlotte.edu,
]

\author[1]{Razvan Bunescu}[%
orcid=0000-0003-2919-3566,
email=razvan.bunescu@charlotte.edu,
url=https://webpages.charlotte.edu/rbunescu,
]
\cormark[1]

\cortext[1]{Corresponding author.}

\begin{abstract}
A restaurant dinner may become a memorable experience due to an unexpected aspect enjoyed by the customer, such as an origami-making station in the waiting area. If aspects that are atypical for a restaurant experience were known in advance, they could be leveraged to make recommendations that have the potential to engender serendipitous experiences, further increasing user satisfaction. Although relatively rare, whenever encountered, atypical aspects often end up being mentioned in reviews due to their memorable quality. Correspondingly, in this paper we introduce the task of detecting atypical aspects in customer reviews. To facilitate the development of extraction models, we manually annotate benchmark datasets of reviews in three domains -- restaurants, hotels, and hair salons, which we use to evaluate a number of language models, ranging from fine-tuning the instruction-based text-to-text transformer Flan-T5 to zero-shot and few-shot prompting of GPT-3.5.
\end{abstract}


\maketitle

\section{Introduction}

\begin{figure*}[t]
\begin{centering}
\includegraphics[width=\textwidth]{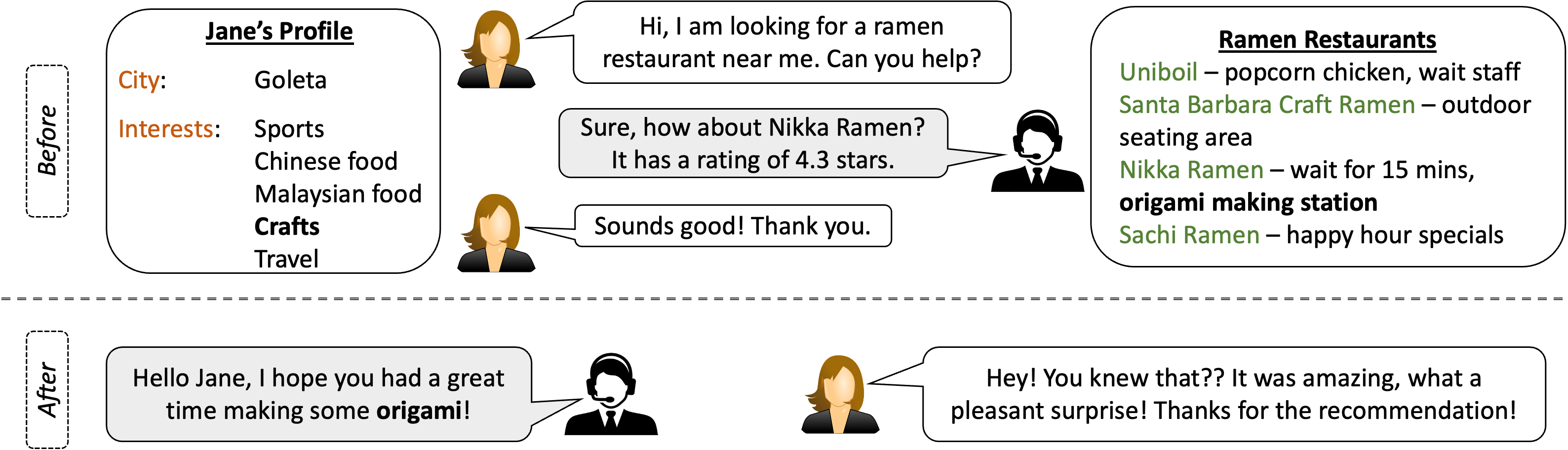}
\par\end{centering}
\label{fig:rec_illustration}
\caption{Example of an interaction with the recommender system showing the potential for serendipity of atypical aspects. The user profile on the left shows a user interest (crafts) that is relevant for the atypical aspect (origami) shown on the right.}
\end{figure*}

When looking for a restaurant or a hotel, people are often faced with an overwhelming number of options matching their search constraints. Even when ranked by their average review scores, there may be numerous high quality choices that satisfy the basic search criteria, especially in a metropolitan area. This may lead to choice overload, or overchoice~\cite{toffler:book70,chernev_choice_2015}, where an individual is presented with a large number of choices that are too difficult to compare, particularly under time constraints \cite{inbar_decision_2011}. Making a decision in the presence of overchoice becomes mentally exhausting and can lead to subsequent impaired self-regulation \cite{vohs_making_2008}, decision paralysis, and anxiety \cite{iyengar_when_2000}. The level of satisfaction that people experience when faced with an increasing number of choices has been observed to follow the well-known Wundt curve \cite{berlyne1973aesthetics}, an inverted U-shape curve originally relating stimulus intensity with its pleasantness. According to this functional dependency, as the number of choices goes up, satisfaction initially increases and then decreases \cite{Shah2007BuyingBA,kaiman:mde18}. In this context, choice overload can be alleviated by reducing the number of consumer choices \cite{schwartz2004paradox} or by making one option stand out and appear better than the others \cite{scheibehenne_can_2010}. To this end, we propose that recommender systems emphasize options that possess aspects with the potential to surprise the user in a positive way, i.e., serendipity. 
Figure~\ref{fig:rec_illustration} illustrates an example where a user Jane is looking for a ramen restaurant in her locality. The system knows that she has been passionate about creating crafts from paper since childhood. Upon being asked for recommendations, the system finds a number of highly rated ramen restaurants, of which Nikita Ramen stands out because it has an origami making station, an atypical aspect for a restaurant, in its waiting area. The system recommends this restaurant to Jane, importantly \textit{without telling her about the origami station}. Upon entering the restaurant, she is very pleasantly surprised to see the origami making station in the waiting area, which brings feelings of nostalgia and happy memories from childhood. She takes some time making various origami figures, before being seated at her table. This serendipitous experience was facilitated by the fact that an origami making station is an atypical aspect for a restaurant, hence it would be experienced as surprising. After the dinner event, the system further confounds her expectations by asking her if she enjoyed the origami making station, which surprises her because she did not expect that the system was responsible for the initial serendipity.

To enable such recommendations with potential for serendipity, the input would ideally consist of three parts:
\begin{enumerate}[leftmargin=*]
    \item The user's query, be it a standalone request or a turn in a longer conversation. This would be used to find the initial, often large set of items that satisfy the user's information need.
    \item The item's data, including not only information about the typical aspects of items from the same category, but also atypical aspects that may generate surprise.
    \item The user's data, especially in terms of their interests, their likes and dislikes. This would be useful for determining if an atypical aspect would be enjoyed by the user, i.e. serendipitous.
\end{enumerate}
To increase the chance of serendipity, the system would need to (i) have knowledge about the user preferences, and (ii) also ensure that the user notices / takes advantage of the atypical aspect, e.g. estimating that there will be some wait involved in the origami example.

In this paper, we introduce a more focused task where we assume that the category of items requested in the user's query is known, e.g. restaurants, and the task is confined to using the item's data to extract aspects that are {\it atypical} for its category, e.g. origami station for restaurants. 
Because users' expectations are shaped by what they think is common for the category of interest in their query, atypical aspects are likely to confound expectations, and hence be perceived as surprising. Henceforth, we will use the term {\it surprising aspects} to refer solely to {\it atypical aspects}.
Whenever atypical aspects are observed for an item, they tend to be more noticeable and often lead to more memorable experiences. As such, they are likely to be mentioned in customer reviews of that item.
Therefore, we use customer reviews as the source of an item's data. At this time, no user data is used as input, which means that the atypical aspects that are extracted, while surprising for the user searching for that particular category of items, cannot be said for sure to lead to serendipity due to unknown user preferences. In short, the task is that of extracting atypical aspects from customer reviews, where atypical is defined to be relative to a predefined item category. To the best of our knowledge, no prior work has looked into extracting atypical aspects from reviews or any other types of item data.

The rest of the paper proceeds as follows. Section~\ref{sec:task} introduces the task of atypical aspect extraction from customer reviews. Section~\ref{sec:datasets} details the development of benchmark datasets of customer reviews that are manually annotated for atypical aspects with respect to three categories: restaurants, hotels, and hair salons. In Section~\ref{sec:approaches} we describe a number of extraction approaches that rely on language models (LM), ranging from Flan-T5 \cite{wei2022finetuned,chung2022scaling} and ChatGPT \cite{chatgpt} in zero-shot or few-shot setting, to fine-tuning of Flan-T5. Experimental evaluations of these models in both extractive and abstractive settings are detailed in Section~\ref{sec:experiments}. The paper ends with related work in Section~\ref{sec:related} and concluding remarks.


\section{Task Definition and Guidelines}
\label{sec:task}

\begin{table*}[t]
    \centering
    \begin{tabular}{p{\textwidth}}
    \toprule
    \noindent $\blacktriangleright$ A group of work friends and I stumbled upon Upper Deck a little over a year ago and everyone from our office has turned Upper Deck into our local watering hole ever since. Their happy hour special is unbeatable, they have a good selection of draft beers, and the food is out of this world good. The stand out feature of Upper Deck is the offering of {\bf life size beer pong} at their outside patio. This takes traditional beer pong and substitutes solo cups with garbage cans (painted to look like solo cups) and Ping Pong balls with dodgeballs They also have a {\bf pool table} and recently added {\bf arcade games} ({\bf nfl blitz 99} beats {\bf madden 15} all day). Get some friends and bring your appetites and some quarters, you won't be disappointed.  \\
    \hdashline[1pt/2.5pt]
    {\it \textbf{The restaurant offers life size beer pong at their outside patio. They have a pool table. They recently added arcade games, such as nfl blitz 99 and madden 15.}} \\
    \midrule
    \noindent $\blacktriangleright$ The big draw of this place is the excellent pizza, which you can have with beer on an outside deck with a view of the {\bf parklands}. It's a nice place to hang out on a sunny afternoon. You can even go for a walk in the {\bf Goose Creek Park} behind the restaurant afterwards to burn off the calories you just consumed. The big minus is that if you go for lunch or in early afternoon, the menu is really limited. This place also seems to attract a goodly number of families with kids at lunch times, probably because it serves pizza and there's a {\bf playground} in the adjoining park, \\
    \hdashline[1pt/2.5pt]
    {\it \textbf{The restaurant has an outside deck with a view of the parklands. Customers can go for a walk in the Goose Creek Park behind the restaurant. There's a playground in the adjoining park.}} \\
    \midrule
    \noindent $\blacktriangleright$ Classic West Philly spot where you can see {\bf local wildlife}. Everyone from {\bf moms} to {\bf anarchists} to {\bf hackers} to {\bf organic gardeners} to {\bf activists} hangs out there. The coffee is excellent, the baked goods are great as well, and if you're working on something, you might run into a possible collaborator there. If you're thinking of moving to West Philly, definitely check out Satellite and the farmer's market. They've replaced the cracked and chipped cups with awesome new cups, which are awesome. \\
    \hdashline[1pt/2.5pt]
    {\it \textbf{In this restaurant you can see local wildlife. Everyone from moms to anarchists to hackers to organic gardeners to activists hangs out there.}}\\
    \midrule
    \noindent $\blacktriangleright$ This is such a cool place! Three words was all it took to add this gem to my list of places to visit while in St. Louis - "{\bf Good Burger Car}!!" YES! They have the car from the movie Good Burger! A movie I was obsessed with as a child \& have since gotten my kids to love just as much! The place is made out of \textcolor{blue}{\textbf{cool, colorful shipping containers}} with many neat decorations, what looks like an \textcolor{blue}{\textbf{alien spaceship}} from Toy Story with a laser on it adorned the top of the place along with a \textcolor{blue}{\textbf{cow}}. Now, on to the food. They have many different options to choose from including create-your-own burgers \& many of their own creations, sandwiches, salads, sides, kid's meals, \& shakes \& floats ... Such a unique place \& worth a visit! They also sell {\bf souvenir T-shirts} \& {\bf hats}, \& my fiance had to get himself a {\bf "HI AF" shirt}. The shirts were heavily influenced by the {\bf movie Good Burger} \& there was one in particular I had my heart set on, unfortunately they did not have it in my size.\\
    \hdashline[1pt/2.5pt]
    {\it \textbf{The restaurant has the car from the movie Good Burger. They sell souvenir T-shirts and hats, and a customer got a "HI AF" shirt. The shirts are heavily influenced by the movie Good Burger.}}\\
    \textcolor{blue}{\textbf{The place is made out of cool, colorful shipping containers, with many neat decorations, what looks like an alien spaceship from Toy Story with a laser on it adorned the top of the place along with a cow.}}\\
    \bottomrule
    \end{tabular}
    \caption{Examples of {\bf extractive} and {\it \textbf{abstractive}} annotations of customer reviews in the restaurants domain. Secondary or optional annotations are shown in \textcolor{blue}{\textbf{blue}}.}
    \label{tab:examples}
\end{table*}

Given a domain category, e.g. {\sc restaurants}, and a customer review of a particular item in that category, e.g. {\it a restaurant}, the task is to extract aspects of that particular item that are {\it atypical of items in its category}. Throughout most of this paper we will use the category of restaurants as an example. All aspects that are related to the core business of a restaurant, including but not limited to food, service, price, opening hours, parking, are considered typical aspects and are not annotated. Conversely, we define and annotate an aspect as atypical if it is not related to the core business of {\sc restaurants}, yet it belongs to or is a feature of {\it the restaurant} ({\sc restaurants} refers generically \cite{carlson_generic_1995} to the restaurant category, whereas {\it the restaurant} refers to a specific restaurant). Correspondingly, in Table~\ref{tab:examples} we show samples taken from 4 reviews, illustrating two types of manual annotations, {\it extractive} and {\it abstractive}, analogous to the extractive \cite{hongyan:sigir99,knight_summarization_2002} and abstractive \cite{jing-2002-using,rush-etal-2015-neural} annotation schemes widely used in summarization datasets\footnote{\url{https://duc.nist.gov}, \url{https://tac.nist.gov}}. A special case is made of aspects related to the ambience or atmosphere of a restaurant: while ambience might be considered as an important part of, and thus subordinated to, the core business of restaurants, there are cases where ambience aspects stand out and become an attraction on their own. When that happens, we annotate them on a secondary, optional layer.

In the extractive annotation, only base noun phrases referring to atypical aspects are annotated. If multiple phrases refer to the same atypical aspects, we only annotate the grounding instance of the coreference chain. For example, the noun phrase "adjoining park" in the second review is not annotated, as it refers to the "Goose Creek Park" which is already annotated. However, if a review mentions an atypical category, such as "arcade games" in the first review or "local wildlife" in the third review, any category instance that is mentioned will also be annotated as atypical, such as "nfl blitz 99" in the first review or "anarchists" in the third review, respectively.
If the atypical aspect is a more complex noun phrase, we only annotate the base noun phrase that expresses the semantic core (often the syntactic head), as in "the {\it Goose Creek Park} behind the restaurant". The extractive annotation is meant to be used together with the original review in downstream applications, which makes it acceptable to annotate only the most important part of the phrase.

In the abstractive annotation, one or more sentences are generated that enumerate the atypical aspects mentioned in the review. The formulation is kept as close as possible to the original text while maintaining naturalness. The generated sentences are intended to be concise, usually maintaining details that are expressed in the same sentence in the review, however keeping out unimportant information about the atypical aspect that is mentioned in other sentences, or details that are vague or uncertain. Sentiment words are maintained only if it helps keep the text natural and faithful to the original. The abstractive annotation is meant to be standalone and used without the original review in downstream tasks, as such it may require some minimal rewriting of the original review formulation, e.g. adding the phrase "the restaurant", or removing opinion words such as "beats" in the first example. Sometimes reviewers use metaphors to refer to atypical aspects, in which case it is important that the abstractive version preserves the metaphorical meaning. This is the case for "local wildlife" in the third review, which refers metaphorically to types of customers that are seen relatively less often in that context.


In many aspect-based sentiment analysis approaches \cite{Semeval-2014}, identifying {\it typical} aspects that are mentioned in a review is done explicitly as {\it aspect term extraction}. However, the task of extracting {\it atypical} aspects, as introduced above, cannot be solved simply by first (a) identifying all typical aspects of restaurants that are mentioned in a review, followed by (b) extracting all other noun phrases, i.e. phrases that do not refer to a typical aspect of a restaurant. In their reviews, people often mention entities or events that are not associated with the restaurant, such as "our office" in the first review, or "the farmer's market" in the third review, and these phrases should not be extracted either. Thus, it is important that the noun phrase refers to an aspect that is {\it associated with the reviewed restaurant}, and that at the same time is atypical of restaurants (or unexpected for a restaurant). Finally, while Table~\ref{tab:examples} may induce the perception that atypical aspects are common, the opposite is actually true. As will be detailed in Section~\ref{sec:datasets} below, it takes going through at least 50 reviews in order to find one review that mentions an atypical aspect. The difficulty of manually finding this "needle in a haystack" further motivates the development of automated approaches for surprising aspect extraction.

\section{Manually Annotated Datasets}
\label{sec:datasets}

We used the Yelp dataset \cite{yelp} as a source of reviews for the 3 target categories: Restaurant ($\sim$5M reviews), Hotel ($\sim$190K reviews), and Hair Salon ($\sim$115K reviews). Because most aspects are expressed as nouns and less frequently as verbs, we use \citet{spacy} to collect lemmas of all nouns and verbs and compute their frequencies for each domain. We rank words in ascending order based on their counts and filter out words that appear with very low frequency, e.g., less than 10 times for the Restaurant domain, as these tend to be spelling mistakes or interjections that are purposely misspelled for extra emphasis, e.g., "amaazzing". We then consider the remaining rare words in ascending order of their frequency as candidate atypical words, extract the reviews that mention them, and read these reviews to determine which occurrences truly refer to an atypical aspect. When reading a review, all atypical aspects are annotated, not only the ones corresponding to the search word. Notwithstanding the heuristic selection of reviews based on the occurrence of rare words, overall this was still a very time-consuming process, because rare words very often appear in a review without necessarily referring to atypical aspects. For example, out of the 43 restaurant reviews that contain the lemma "poncho", in only 1 review the word "ponchos" was deemed to refer to an atypical aspect (the restaurant was selling them). The other 42 reviews contained references to ponchos that were not associated with the restaurants itself, e.g. staff helping customers put their ponchos on a rainy day, or customers describing their arrival at the restaurant on a rainy day. As we went down the list of rare words, their frequency increased, resulting in a larger number of reviews to skim through for each rare word. Overall, for the Restaurant dataset, we used as search words the rare words that appeared with a frequency of up to 187. Upon semi-automatically sifting through the $\sim$97K reviews found to contain these words, we were able to collect 114 reviews that contained atypical aspects. On average, one hour of following this process led to finding between 2 and 3 reviews containing atypical aspects for the restaurant category, whereas for the hair salon category it took on average two hours to find 1 atypical reviews. Henceforth the term {\it atypical review} will be used to refer to a review that contains one or more atypical aspects; analogously, the term {\it typical review} will be used to refer to reviews that do not contain any atypical aspect.

As illustrated in the examples from Section~\ref{sec:task}, we organize annotations of atypical aspects on two layers:
\begin{itemize}[leftmargin=*]
    \item A {\it primary} layer that contains atypical aspects that are clearly not connected to any core feature of that domain.
    \item A {\it secondary} layer that contains atypical aspects that are related to a typical aspect, such as ambiance or location, but that stand out and are interesting on their own, separate from the core features of the domain.
\end{itemize}
For example, \textit{'I was even encouraged to visit their \textbf{petting zoo} in the back'} would be considered a primary atypical aspect in any of the 3 categories, whereas \textit{'There is an \textbf{interesting giant stuffed spider} that goes up and down when the door leading to the bathrooms opens and closes'} would be annotated as a secondary atypical aspect.

Table~\ref{tab:dataset} shows summary statistics for the 3 datasets, one for each domain (category), split between data used for training and testing, and data used for development. Under the Primary column, we show the number of atypical reviews and atypical aspect annotations contained in them. The next column shows the same statistics for when both primary and secondary atypical aspects are considered. The total number of reviews in each dataset, shown in the third column, is about double the number of primary atypical reviews, reflecting a balanced dataset where the number of typical reviews was selected to be about the same as the number of  atypical reviews.

We computed inter-annotator agreement (ITA) on both the extractive and abstractive annotations in the development sets of the Restaurant and Hair Salon domains. The ITA metrics are shown in Table~\ref{tab:evaluation} and are calculated by assuming one annotator provides the ground truth while the other annotator is considered as the system.

\begin{table*}[t]
\centering
 \caption{Statistics for the 3 datasets, split between Train+Test and Development (Dev). The number of atypical reviews and atypical aspects are presented separately for primary (Primary) vs. both primary and secondary atypical aspects ($+$ Secondary).}
\begin{tabular}{@{}ccccccc@{}}
\toprule
\multicolumn{1}{l}{\textbf{Domain}} & \multicolumn{1}{l}{\textbf{Dataset split}} & \multicolumn{2}{c}{\begin{tabular}{c} \textbf{Primary}\end{tabular}} & \multicolumn{2}{c}{\begin{tabular}{c} $+$ \textbf{Secondary}\end{tabular}} & \multicolumn{1}{c}{\begin{tabular}{c} \textbf{Total}\end{tabular}}  \\
 \multicolumn{1}{l}{} & \multicolumn{1}{l}{} & \multicolumn{1}{c}{\# reviews} & \multicolumn{1}{c}{\# aspects} & \multicolumn{1}{c}{\# reviews} & \multicolumn{1}{c}{\# aspects} & \multicolumn{1}{c}{\textbf{reviews}}\\
\midrule
\multirow{2}{*}{Restaurant} & Train+Test & 100 & 253 & 107 & 340 & 200 \\
 & Dev & 14 & 32 & 16 & 46 & 28 \\
 \cmidrule(lr){2-7}
\multirow{2}{*}{Hotel} & Train+Test & 69 & 274 & 85 & 401 & 150 \\
 & Dev & 10 & 33 & 11 & 49 & 20\\
\cmidrule(lr){2-7}
\multirow{2}{*}{Hair Salon} & Train+Test & 45 & 147 & 48 & 181 &  90 \\
 & Dev & 5 & 24 & 5 & 29 & 10 \\
\bottomrule
\end{tabular}
\label{tab:dataset}
\end{table*}

\section{LM-based Approaches}
\label{sec:approaches}


This section describes our Language Model (LM) based approaches to detecting surprising aspects in customer reviews. We experimented with 2 language models: Flan-T5 and ChatGPT. The 3 billion parameter FLAN-T5 is an encoder-decoder transformer based on the T5 model~\cite{raffel2020exploring} that was further instruction-tuned on the FLAN dataset~\cite{wei2022finetuned,chung2022scaling}. We decided to use the FLAN-T5 model due to its exposure to narratives in the style of reviews, e.g., blog posts, during pre-training on the C4 corpus~\cite{raffel2020exploring}, and also due to its instruction-tuning on summarization and sentiment analysis tasks. We also experimented with zero-shot and few-shot prompting of the much larger ChatGPT (\texttt{gpt-3.5})~\cite{chatgpt} in order to evaluate the performance of a state-of-the-art language model without any fine tuning.

LMs are known to be sensitive to word choices \cite{webson-2022-prompt}.
In the case of zero-shot/few-shot experiments, we tried between 10 and 20 different prompts for each type of annotation and selected the ones that performed the best on the development data. We noticed that, given the description of the task along with examples of typical aspects in the prompt helped model to identify the multiple atypical aspects present in a review. We also observed that the more examples of typical aspects were given in the prompt the easier it was for the model to identify atypical aspects. We also had to try multiple prompt formulations to instruct the LM to align with an expected output format. For 5-shot experiments, various sets of 5 examples from the development set were first selected at random, then manually checked for diversity in terms of difficulty, number, and types of atypical aspects. Of these, we selected the 5-shot examples that yielded the highest  performance on the development set.

With the exception of abstractive generation for Hotels, which did not benefit from a prompt, the FLAN-T5 fine-tuning experiments employed the following prompts:
\begin{itemize}[leftmargin=*]
    \item[$\blacktriangleright$] \textbf{Fine-tuning FLAN-T5 Extractive Prompt}: {\it question: Based on the following restaurant review, list aspects that are atypical for a restaurant. Separate them using commas. context:} \{\{\texttt{Review}\}\}
    \item[$\blacktriangleright$] \textbf{Fine-tuning FLAN-T5 Abstractive Prompt}: {\it question: Based on the following restaurant review, what are the atypical aspects for a restaurant? context:} \{\{\texttt{Review}\}\}
\end{itemize}

In the 0-shot setup for ChatGPT, we include an instruction to either extract lists of atypical aspects (extractive) or to generate naturally sounding text about the atypical aspects in the review (abstractive):
\begin{itemize}[leftmargin=*]
    \item[$\blacktriangleright$] \textbf{0-shot ChatGPT Extractive Prompt}: {\it Given the following restaurant review, can you list atypical aspects for a restaurant? Atypical aspects are not related to service, food, drinks, location, price, menu, discounts, policies, staff, customer satisfaction, or other items commonly associated with a restaurant. Please be precise in your response; it should contain only atypical aspects associated with the restaurant that is reviewed. Extract base noun phrases in the output format: 'Atypical aspects: aspect 1, aspect 2, aspect 3.' Output} \texttt{$\langle$None$\langle$} {\it if there are no atypical aspects. Please follow the output format strictly.}\\
    {\it Passage: } \{\{\texttt{Review}\}\}
    \item[$\blacktriangleright$] \textbf{0-shot ChatGPT Abstractive Prompt}: {\it Which aspects mentioned in the review are atypical for a restaurant? Unlike common aspects such as service, food, drinks, location, price, menu, discounts, policies, staff, or customer satisfaction, atypical aspects are not commonly associated with a restaurant. In the output, formulate each aspect as sentences, e.g., "Atypical aspects:} -- {\it The restaurant has $\langle$aspect 1$\rangle$.} -- {\it The restaurant has $\langle$aspect 2$\rangle$.} -- {\it The restaurant has $\langle$aspect 3$\rangle$."}
    {\it If there are no atypical aspects, output "None".}\\
    {\it Passage: } \{\{\texttt{Review}\}\}
\end{itemize}
In the few-shot setup for ChatGPT, we include in the prompt both the instruction and 5 worked-out examples:
\begin{itemize}[leftmargin=*]
    \item[$\blacktriangleright$] \textbf{Few-shot ChatGPT Extractive and Abstractive Prompt}: {\it Given the following restaurant review, can you list atypical aspects for a restaurant? Atypical aspects are not related to service, food, drinks, location, price, menu, discounts, policies, staff, customer satisfaction or other types of items that are commonly associated with a restaurant. Please be precise in your response, which should contain only atypical aspects that are associated with the restaurant that is reviewed. Output } \texttt{<None>} {\it if there are no atypical aspects.}
    \begin{itemize}[leftmargin=*]
        \item[] {\it Example 1:} \{\{\texttt{Example Review 1}\}\} {\it Atypical aspects:} \{\{comma-separated extractive annotations OR bullet-listed abstractive sentences\}\} ... {\it Examples 2, 3, 4, 5}
        \item[] {\it Can you try for the review below?} 
        \{\{\texttt{Review}\}\}
    \end{itemize}
\end{itemize}

We use the Hugging Face Transformers package~\cite{wolf2020transformers} for fine-tuning Flan-T5 with the following hyper-parameters: an effective batch size of 32, a number of epochs of 30, a learning rate of 3e-5 for Restaurants and 5e-5 for Hotels and Hair Salons, a weight decay of 0.001, and a generation max length set to 512. Those hyper-parameter values were found through tuning on the development portion of each dataset. We perform the fine-tuning experiments on a high-performance computing cluster using $8$ CPU cores, $128$ GB RAM, and 2 A100 80 GB GPUs, for around 96 hours. We use the OpenAI API Python package \cite{openai_api} for ChatGPT, where we do greedy decoding by setting the temperature parameter to 0.


\section{Experimental Evaluations}
\label{sec:experiments}

\begin{table*}[t]
    \centering
     \caption{Extractive (exact vs. partial match) and abstractive results (\%), on primary (default) and primary + secondary atypical aspects across the 3 domains. Precision (P), Recall (R), and F1 are reported for the extractive setting. F1 for Rouge-1 (R-1), Rouge-2 (R-2), rougeLsum (RLS) and BERTScore is reported for the abstractive setting. Best results in each domain are in \textbf{bold}.}
     
    \begin{tabular}{lrrr|rrr|rrrr}
        \toprule
         \multicolumn{1}{l}{\textbf{Language Model \&}} & \multicolumn{3}{c}{\begin{tabular}{c} \textbf{Exact} \textbf{Match}\end{tabular}} & \multicolumn{3}{c}{\begin{tabular}{c} \textbf{Partial Match}\end{tabular}} & \multicolumn{4}{c}{\begin{tabular}{c} \textbf{Abstractive}\end{tabular}}  \\
         \multicolumn{1}{l}{\textbf{Experimental Setup}} &  \multicolumn{1}{c}{P} & \multicolumn{1}{c}{R} & \multicolumn{1}{c}{F1} & \multicolumn{1}{c}{P} & \multicolumn{1}{c}{R} & \multicolumn{1}{c}{F1} & \multicolumn{1}{c}{R-1} & \multicolumn{1}{c}{R-2} & \multicolumn{1}{c}{RLS} & \multicolumn{1}{c}{BERT}\\
        \midrule
        & \multicolumn{1}{l}{} & \multicolumn{1}{l}{} & \multicolumn{1}{l}{} & \multicolumn{3}{c}{\textbf{Restaurants}} & & \\
     
        ChatGPT (0-shot) & 22.0 & 37.7 & 27.7 & 27.9 & 56.8 & 37.4 & 35.0 & 24.0 & 52.0 & 66.0 \\
        \hspace{1.15cm}  + secondary & 25.1 & 35.2 & 29.3 & 32.5 & 56.0 & 41.2 & 42.0 & 29.0 & 53.0 & 68.0 \\
        ChatGPT (5-shot) & 33.0 & 43.6 & 37.6 & 39.3 & 61.7 & 48.0 & 42.0 & 31.0 & 59.0 & 71.0\\
        \hspace{1.15cm}  + secondary & 26.8 & 38.9 & 31.7 & 33.5 & 60.7 & 43.2 & 46.0 & 36.0 & 57.0 & 70.0\\
        FLAN-T5 (0-shot) & 28.9 & 23.5 & 25.9 & 32.4 & 28.7 & 30.5 & 32.0 & 24.0 & 44.0 & 54.0\\
        \hspace{1.15cm}  + secondary & 27.9 & 18.5 & 22.3 & 32.3 & 23.9 & 27.4 & 30.0 & 22.0 & 44.0 & 56.0\\
        FLAN-T5 (fine-tuned) & \textbf{67.5} & \textbf{60.2} & \textbf{63.4} & \textbf{72.9} & \textbf{65.3} & \textbf{68.6} & \textbf{58.0} & \textbf{50.0} & \textbf{73.0} & \textbf{79.0}\\
        \hspace{1.15cm}  + secondary & 57.7 & 58.6 & 56.4 & 64.2 & 64.4 & 62.3 & 56.0 & 48.0 & 68.0 & 75.0\\
        Human ITA (est.) & 80.0 & 78.26 & 79.12 & 85.56 & 82.61 & 84.06 & 65.0 & 51.0 & 75.0 & 80.0 \\
        \hspace{1.15cm}  + secondary &  75.4 & 79.3 & 77.3 & 82.4 & 86.8 & 84.5 & 71.0 & 54.0 & 76.0 & 86.0 \\
        & \multicolumn{1}{l}{} & \multicolumn{1}{l}{} & \multicolumn{1}{l}{} & \multicolumn{3}{c}{\textbf{Hotels}} & & & \\
    
        ChatGPT (5-shot) & 30.5 & 35.2 & 32.7 & 34.8 & 46.9 & 39.9 & \textbf{34.0} & \textbf{26.0} & 56.0 & \textbf{65.0} \\
        FLAN-T5 (fine-tuned) & \textbf{60.2} & \textbf{54.9} & \textbf{55.9} & \textbf{63.8} & \textbf{57.6} & \textbf{59.0 }& {\bf 34.0} & {\bf 26.0} & \textbf{59.0} & 63.0 \\

        & \multicolumn{1}{l}{} & \multicolumn{1}{l}{} & \multicolumn{1}{l}{} & \multicolumn{3}{c}{\textbf{Hair Salons}} & & &\\

        ChatGPT (5-shot) & 34.7 & 43.2 & 38.5 & 42.4 & 61.1 & 50.1 & \textbf{48.0} & \textbf{37.0} & 57.0 & \textbf{69.0} \\
        FLAN-T5 (fine-tuned) & \textbf{66.6} & \textbf{62.9} & \textbf{63.9} & \textbf{75.1} & \textbf{69.1 }& \textbf{71.1} & 38.0 & 33.0 & \textbf{61.0} & 65.0 \\
        Human ITA (est.) & 92.3 & 82.8 & 87.3 & 94.2 & 85.1 & 89.4 & 59.0 & 41.0 & 78.0 & 91.0 \\
        \hspace{1.15cm}  + secondary &  74.2 & 67.6 & 70.8 & 79.0 & 71.8 & 75.3 & 65.0 & 47.0 & 77.0 & 91.0 \\
  
        \bottomrule
    \end{tabular}
   
    \label{tab:evaluation}
\end{table*}

The LM-based approaches are evaluated in a 10-fold scenario where the Train+Test reviews dataset is partitioned into 10 folds, 9 folds are used for training and 1 fold is used for testing. This process is repeated 10 times until each fold in the dataset is used as a test fold. The metrics computed across the 10 folds are then micro-averaged yielding the final evaluation metric. For the extractive evaluation, we report the precision, recall, and $F_1$ scores for the {\bf exact} and {\bf partial} matches of the extracted base noun phrase (BNP) with the ground truth phrase.

In the exact match method, an extracted phrase is considered correct if and only if it matches exactly a ground truth (gold) phrase. Precision (P) and Recall (R) are then computed as follows:
\begin{itemize}
    \item[] $P = \# \mbox{\it{correct extracted BNPs}} \; / \; \# \mbox{\it{extracted BNPs}}$
    \item[] $R = \# \mbox{\it{correct extracted BNPs}} \; / \; \# \mbox{\it{gold BNPs}}$
\end{itemize}

In the partial match method, we use the greedy method shown in Algorithm~\ref{alg:cap} to compute a bipartite matching between gold phrases $gp$ and extracted phrases $ep$ that aimed at maximizing their word overlap $|ep \cap gp|$ in total. The overlaps between extracted and gold phrases are then used to compute precision, recall, and $F_1$.


\begin{algorithm}
   \SetKwInOut{KwIn}{Input}
   \SetKwInOut{KwOut}{Output}

   \KwIn{Gold Phrases $GP$, Extracted Phrases $EP$}
   \KwOut{Precision (P), Recall (R), and $F_1$}
    \tcp{$TPe$ = \# True Positives w.r.t $EP$}
    \tcp{$TPg$ = \# True Positives w.r.t $GP$}
    \tcp{$FP$ = \# False Positives}
    \tcp{$FN$ = \# False Negatives}
   \For{gp in GP}{
       \eIf{EP is empty}{
           $FN \gets FN + |gp| / |gp|$
        }{ 
        Find $ep \in EP$ that has maximum Jaccard similarity with $gp$\\
            $TPe \gets TPe + |ep \cap gp| / |ep|$ \\
            $TPg \gets TPg + |ep \cap gp| / |gp|$ \\
            $FP \gets FP + |ep - gp| / |ep|$ \\
            $FN \gets FN + |gp - ep| / |gp|$ \\
        }
        Remove $ep$ from the set $EP$
    }
    \For{ep in EP}{
        $FP \gets FP + |ep| / |ep|$
    }
    $P \gets TPe / (TPe + FP)$, \ \ $R \gets TPg / (TPg + FN)$ \\
    \KwRet{$P, \; R, \; F_1 \gets 2 P R / (P + R)$}
    \caption{PartialMatchMetrics($GP$, $EP$)}\label{alg:cap}
 \end{algorithm}


For the abstractive evaluation, we follow prior work in summarization \cite{tangsali-etal-2022-abstractive, ahuja-etal-2022-aspectnews, mrini-etal-2021-rewards} and compare the generated output with the ground truth using BERT F1 Score~\cite{bertscore} instantiated with DeBERTa~\cite{he2020deberta}, Rouge-1, Rouge-2, and Rouge-L-Sum~\cite{lin-2004-rouge}. In the case of typical reviews, the prediction, i.e. the generated text, should be empty, reflecting no atypical aspect. When using RougeLsum and BERTScore to evaluate the model output on a typical review, $F_1$ is calculated as 1.0 if the model generates an empty text, and 0.0 if it generates a non-empty text.

The overall experimental results are shown in Table~\ref{tab:evaluation}. For Restaurants, we show results on extracting primary atypical aspects as well as results on extracting both primary and secondary atypical aspects. Since fine-tuned Flan-T5 and ChatGPT (5-shot) obtained the best results on Restaurants, they were selected to be evaluated on the other two domains, using solely primary atypical aspects. Fine-tuning FLAN-T5 yields the best performance in the extractive task across all domains. While we observe a big performance gap between ChatGPT and fine-tuned FLAN-T5 in the extractive setting, that gap shrinks considerably in the abstractive setting for Hotels and Hair Salons, where ChatGPT (5-shot) occasionally outperforms FLAN-T5 on some of the metrics. For both LMs, the Hotel domain appears to be more challenging. Compared to the other domains, atypical aspects are more common and more diverse in hotels, likely because hotels try to differentiate themselves from other hotels more than restaurants or hair salons do. Table~\ref{tab:dataset} shows that indeed there are more primary and secondary atypical aspects per review in the hotel domain.

To determine how well Flan-T5 generalizes to unseen atypical aspects in the Restaurant domain, we manually created groupings of atypical aspects where semantically similar atypical aspects, e.g. greeting cards and anniversary gifts, are grouped together, such that aspects in different groups are semantically very different. We then partition the set of groups into 10 folds of groups, which ensures that the atypical aspects that the language model sees in the test fold have not been seen during training (either literally or semantically similar). Upon fine-tuning and evaluating FLAN-T5 on this dataset, we observe a similar precision as reported in Table~\ref{tab:evaluation}, however, there is a significant drop in recall from 60.2 to 46.1 for primary atypical aspects and from 58.6 to 49.3 when extracting both primary and secondary atypical aspects. Improving generalization to semantically novel atypical aspects is therefore an interesting avenue for future work.

To get a sense of the real performance in the abstractive setting for primary aspects, we also performed a manual evaluation of the fine-tuned Flan-T5 and 5-shot ChatGPT outputs. To enable calculation of precision and recall, we manually label and count the following: (A) Atypical aspects from the model output that are semantically similar to gold aspects; (B) Gold aspects that are missing from the model output; (C) Typical aspects or other types of entities that should not have been extracted as atypical; (D) Extra details about atypical aspects; (E) Secondary atypical aspects; (F) Typical reviews for which the model correctly generates an empty string.
The first 4 types are illustrated on the example below, where A = 2, B = 1, C = 1, D = 1, and E = 1:
\begin{itemize}[leftmargin=*]
    \item Gold = ["On the weekends kids can experience 'Mark the Balloon Guy' at the restaurant. They have stuffed animal/puppets for sale at the front.\textsuperscript{B}", "The restaurant has therapeutic sketching at every table."]
    \item Output = ["The restaurant has a balloon artist on weekends.\textsuperscript{A} The restaurant has traditional wooden baseball stadium seats for waiting.\textsuperscript{E}", "The restaurant offers therapeutic sketching at every table.\textsuperscript{A} If your sketch is good, it will be on the wall.\textsuperscript{D} The restaurant has limited menu options before 4:30 pm.\textsuperscript{C}"]
\end{itemize}
These counts are then used to compute a lenient precision as $P = (A + F) / (A + F + C)$ and a strict precision as $P = (A + F) / (A + F + C + D + E)$, whereas recall is computed as $R = (A + F) / (A + F + B)$.
Correspondingly, the fine-tuned Flan-T5 obtains a lenient and strict $F_1$ of 77.5\% and 76.5\%, respectively, whereas ChatGPT obtains a lenient and strict $F_1$ of 71.3\% and 65.9\%, respectively. Looking at Table~\ref{tab:evaluation}, the strict $F_1$ is in between RLS and BERT, further supporting the use of these automated metrics. Overall, manual evaluation shows that models perform better in the abstractive vs. extractive setting.


        


Error analysis reveals that fine-tuned Flan-T5 is more succinct in its answers, leading it to sometimes ignore atypical aspects in its response. Conversely, ChatGPT tends to be more verbose, often generating unnecessary details about the atypical aspects that it extracts, or mistaking typical for atypical aspects. For instance, ChatGPT extracted a food-related aspect in the second sentence in {\it "The restaurant has an annual Valentine's Date Night with table service, a caricature artist, and a piano player. \textcolor{red}{The restaurant offers in-house made chocolate-covered strawberries.}"}. ChatGPT also tends to extract one-time customer experiences which are not atypical aspects of a restaurant, as shown in the second sentence in {\it "The restaurant has a sister restaurant next door with a lively fun band. \textcolor{red}{The server, Peter, gave very helpful suggestions.}"}. On the other hand, fine-tuned FLAN-T5 mistakenly considers {\it "\textcolor{red}{farmer's market}"} to be an atypical aspect in the third review from Table~\ref{tab:examples}. Also, when the sole atypical aspect of a business is in its proximity, the fine-tuned FLAN-T5 tends to not extract it, as in the model failing to generate the ground truth abstractive sentences {\it "\textcolor{violet}{They are close to a dock area where customers can board paddle cruises.}" , "\textcolor{violet}{Customers can catch an airboat ride down the road.}"}

\section{Related Work}
\label{sec:related}


Addressing overchoice is a core focus of recommender systems and it is typically addressed by recommending between 5 to 20 attractive and diverse items~\cite{bollen_overchoice_2010}, based on user preferences~\cite{jakob_reviewrec_2009, zhang_reviewrec_2014}, user ratings, item attributes, or user reviews~\cite{musto_absarec_2017, selmene_sentanalysis_2020, li_latentfeat_2022}. Recommending items with potential for serendipity is one way of diversifying an item set. In \cite{panagiotis_unex_2011}, unexpectedness is defined as the distance of an item from a set of obvious items for that user, relative to the user's preferred level of unexpectedness. \citet{li_unexrec_2020} recommended unexpected items by modeling user interests as clusters of historical data in a latent space and calculating the weighted distance between a new item and the clusters of interests. \citet{kotkov_serDataset_2018} crowd-sourced serendipity labels for a movie dataset using multiple definitions of serendipity. 

User reviews have been used to learn latent features of users~\cite{li_latentfeat_2022}, extract sentiment~\cite{selmene_sentanalysis_2020}, derive user preferences~\cite{zhang_reviewrec_2014}, or to perform aspect-based sentiment analysis in order to recommend better quality products with aspects relevant to the user~\cite{musto_absarec_2017}.
Conversational recommender systems use reviews to provide explanations \cite{musto_justify_2019}, to maintain fluency in conversation \cite{lu-cres-2021}, or to understand the user's requirements by asking questions about aspects mentioned in reviews \cite{zhang_cres_2018}.
 %

Aspect-based sentiment analysis (ABSA) is a technique that aims to extract topic-specific aspects mentioned in a review, together with any associated sentiment.
There are different approaches to extracting aspect terms, with some focusing on using term frequency in the corpus~\cite{bauman_abr_2017, souza_absa_2022} or using semantic clustering of prominent aspects~\cite{luo-etal-2018-extra}. 
\citet{qiu_opinionext_2011} only extract aspects that have an expressed opinion, using expanded opinion lexicons containing adjectives. As detailed at the end of Section~\ref{sec:task}, extracting atypical aspects cannot be addressed simply as the logical complement of ABSA.


Prior work investigated using Language Models (LMs) in recommender systems. \citet{Zhang2021} investigate using LMs as a recommender system by formulating a movie recommendation task as a multi-token cloze task and find that LMs underperform traditional recommender systems such as GRU4Rec in both zero-shot and fine-tuned settings. LMs have also been used to elicit user preferences in natural language given historical interactions and prior selections for more personalized recommendations~\cite{chen2023palr}. Such a model can benefit a complete implementation of the system illustrated in Figure~\ref{fig:rec_illustration}, where it would automatically infer Jane's interests. Furthermore, LMs have been used for explaining why a specific recommendation was made to the user~\cite{gao2023chat}.

Atypical aspects are valuable due to their potential to create serendipity, i.e. experiences that are both unexpected (surprising) and relevant (positive) for the user. Unlike \citet{kotkov_serDataset_2018}, we do not consider novelty to be a required dimension of serendipity. Finally, surprise can appear from other sources, be it the experience of a typical aspect that stands out (such as a unique dish offered by a restaurant), or the accidental discovery of a restaurant as shown in the first sentence in Table~\ref{tab:examples} and studied in \cite{xi_2023_serendipity}. All of these different sources of surprise have great potential for setting up serendipity.

\section{Conclusion and Future Work}

We introduced the new task of extracting atypical aspects from customer reviews. 
To enable training and evaluation of atypical aspect extraction models, we manually annotated two layers of atypical aspects in customer reviews from three domains. While experimental evaluations using few-shot prompting of ChatGPT and fine-tuning of Flan-T5 show promising results, there is still a substantial gap relative to human performance, as shown by the higher ITA. Future work includes enhancing reproducibility by using open-source LMs \cite{liesenfeld:cui23}, prototyping a recommender system that leverages atypical aspects, and a user study verifying their utility.

To facilitate reproducibility and future progress, we make the code and the datasets publicly available at \url{https://github.com/smitanannaware/XtrAtA}.





\begin{acknowledgments}
This research was partly supported by the United States Air Force (USAF) under Contract No. FA8750-21-C-0075. 
\end{acknowledgments}

\bibliography{kars-recsys23}

\appendix

\end{document}